\pdfoutput=1

\documentclass[11pt]{article}

\usepackage[]{acl}

\usepackage{times}
\usepackage{latexsym}

\usepackage[T1]{fontenc}

\usepackage[utf8]{inputenc}

\usepackage{microtype}

\usepackage{subcaption}
\usepackage{graphicx}
\usepackage{booktabs}
\usepackage{multicol}
\usepackage{multirow}
\usepackage{amsmath}
\usepackage{amssymb}
\usepackage{bbding}

\usepackage{algorithm}  
\usepackage{algpseudocode}  
\usepackage{amsmath}

\usepackage{cleveref}
\crefname{section}{§}{§§}
\Crefname{section}{§}{§§}

\usepackage{multirow}
\usepackage{amsmath}
\usepackage{capt-of}
\usepackage{tabularx}
\usepackage{epsfig}
\usepackage{amssymb}
\usepackage{amsfonts}
\usepackage{booktabs}
\usepackage{scalerel}
\usepackage[inline]{enumitem}
\usepackage{listings}
\usepackage{varwidth}
\usepackage[export]{adjustbox}
\usepackage{tikz}
\usetikzlibrary{tikzmark}
\usepackage{cleveref}

\usepackage{stmaryrd}
\usepackage{bbm}

\usepackage{algorithm}

\definecolor{deepblue}{rgb}{0,0,0.5}
\definecolor{officeblue}{RGB}{0,102,204}
\definecolor{deepred}{rgb}{0.6,0,0}
\definecolor{deepgreen}{rgb}{0,0.5,0}
\definecolor{mybrickred}{RGB}{182,50,28}

\definecolor{fillcolor}{RGB}{216,217,252}

\renewcommand{\algorithmicrequire}{\textbf{Input:}}
\algnewcommand\algorithmicrequireb{{\hspace{0.85cm}}}
\algnewcommand\INPTDESCB{\item[\algorithmicrequireb]}
\renewcommand{\algorithmicensure}{\textbf{Output:}}
\algnewcommand\algorithmicfuncdesc{\textbf{Function:}}
\algnewcommand\FUNCDESC{\item[\algorithmicfuncdesc]}
\algnewcommand\algorithmicfuncdescb{{\hspace{1.48cm}}}
\algnewcommand\FUNCDESCB{\item[\algorithmicfuncdescb]}
\algnewcommand{\algorithmicgoto}{\textbf{goto}}
\algnewcommand{\Goto}[1]{\algorithmicgoto~\ref{#1}}

\usepackage{amsmath,amsfonts,bm}

\def\eqref#1{equation~\ref{#1}}

\def\1{\bm{1}}

\DeclareMathAlphabet{\mathsfit}{\encodingdefault}{\sfdefault}{m}{sl}
\SetMathAlphabet{\mathsfit}{bold}{\encodingdefault}{\sfdefault}{bx}{n}

\title{CLIP Models are Few-shot Learners: \\ Empirical Studies on VQA and Visual Entailment}

\author{Haoyu Song$^{\dagger}$\thanks{\ \ Contribution during internship at Microsoft Research.}, \ Li Dong$^{\ddagger}$, \ Wei-Nan Zhang$^{\dagger}$, \ Ting Liu$^{\dagger}$, \ Furu Wei$^{\ddagger}$ \\
        $^{\dagger}$ Harbin Institute of Technology \\  $^{\ddagger}$ Microsoft Research \\
        \texttt{\{hysong,wnzhang,tliu\}@ir.hit.edu.cn} \\ \texttt{\{lidong1,fuwei\}@microsoft.com}}

\begin{document}

\maketitle

\begin{abstract}

CLIP has shown a remarkable zero-shot capability on a wide range of vision tasks. Previously, CLIP is only regarded as a powerful visual encoder. However, after being pre-trained by language supervision from a large amount of image-caption pairs, CLIP itself should also have acquired some few-shot abilities for vision-language tasks. In this work, we empirically show that CLIP can be a strong vision-language few-shot learner by leveraging the power of language. We first evaluate CLIP's zero-shot performance on a typical visual question answering task and demonstrate a zero-shot cross-modality transfer capability of CLIP on the visual entailment task. Then we propose a parameter-efficient fine-tuning strategy to boost the few-shot performance on the vqa task.
 We achieve competitive zero/few-shot results on the visual question answering and visual entailment tasks without introducing any additional pre-training procedure.

\end{abstract}

\section{Introduction}

Vision-language understanding (VLU) tasks, such as visual question answering~\cite{antol2015vqa} and visual entailment~\cite{xie2019visual}, test a system's ability to comprehensively understand the semantics of both visual world and natural language.
To capture the alignment between vision and language, various efforts have been made to build the vision-language pre-trained models~\cite{lu2019vilbert,chen2020uniter,su2020vl,zhang2021vinvl,wang2021vlmo}. Despite their superior performances, these methods have extensively utilized human-annotated training data that are expensive or require expert knowledge, such as object detection datasets~\cite{lin2014microsoft,kuznetsova2020open} and aligned image-text pairs~\cite{deng2009imagenet,sharma2018conceptual}.
Collecting such datasets requires heavy work on data gathering and human annotation, and thus their scales are only in the realm of tens of millions, which are much smaller than the Internet text corpora for NLP pre-training~\cite{devlin2019bert,brown2020language}.

\begin{figure}[t]
\centering
\includegraphics[width=.99\columnwidth]{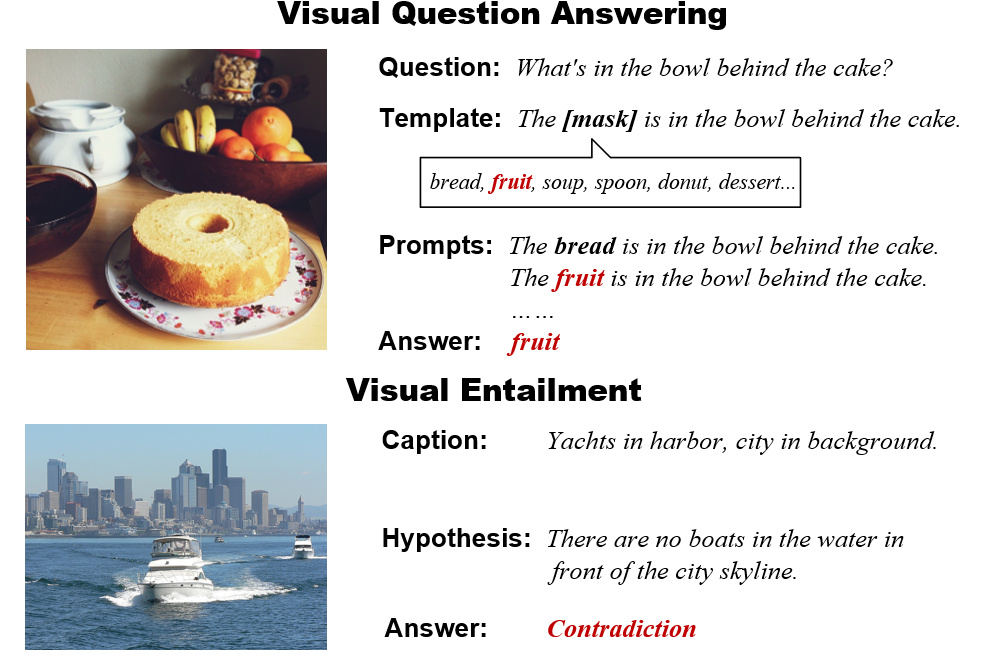}
\caption{ Examples of the two vision-language understanding tasks. For VQA, language prompts are used. For visual entailment, caption and hypothesis, i.e., text-text pairs, are used in training, while image and hypothesis, i.e., image-text pairs, are used at inference. }
\label{fig:1}
\end{figure}

Recently, CLIP~\cite{radford2021learning} has been proposed to learn visual concepts with natural language supervision, where its 400 million image-text pairs are crawled from the Internet. CLIP consists of a visual encoder and a text encoder, and it learns visual representations by aligning images and texts through contrastive loss. In this way, CLIP achieves strong zero-shot performances on vision benchmarks such as ImageNet. Besides,~\citet{shen2021much} prove that CLIP could be leveraged as a strong visual encoder to benefit downstream vision-language tasks.
However, there are two major differences between CLIP and previous visual encoders: 1) it is trained on much larger yet noisy web data, and 2) it has a shallow interaction between vision and language. The first feature promises the generalization ability of CLIP, and the second one equips alignment ability across modalities. Could the strong zero-shot ability of CLIP be transferred to vision-language understanding tasks?

To answer the above question, in this work, we empirically study how to transfer CLIP's zero-shot ability into VLU tasks and further turn CLIP into a few-shot learner. We carried out experiments on two VLU tasks: 1) visual question answering, where the model needs to give an answer according to the details of an image and a natural sentence question, and 2) visual entailment, where the model needs to determine the entailment relation between an image and a natural sentence. Figure~\ref{fig:1} demonstrates the basic forms of the two studied tasks.

For the zero-shot visual question answering task, the key to a successful zero-shot capability transfer is to mitigate the gap between the pre-training task of CLIP and the task form of question answering. Inspired by the recent advancements of few-shot learning in NLP~\cite{schick2021s,gao-etal-2021-making}, we address this issue by introducing a two-step prompt generation strategy, including automatic conversions from question to statement to get masked templates, and a span-infilling with generative pre-trained T5 model~\cite{raffel2020exploring} to get candidate answers. 

We explore a zero-shot cross-modality (language and vision) transfer capability through the visual entailment task. Specifically, we replace the image with its captions during training and only update a small classification layer. Then at inference, as usual, we still use image-text pairs for testing. This allows us to investigate how well the language and vision representations are aligned in CLIP models.

We further leverage few-shot learning to improve CLIP's visual question answering performance based on the zero-shot transferring methods. We find that optimizing only bias and normalization (BiNor) parameters would make better use of limited examples and yield better results than the latest few-shot model \textit{Frozen}~\cite{tsimpoukelli2021multimodal}. Experiments confirm that CLIP models can be good vision-language few-shot learners.

Our contributions are summarized as follows:
\begin{itemize}
  \item To the best of our knowledge, this is the first work that studies how to transfer CLIP's zero-shot capabilities into VLU tasks and confirms CLIP models can be good few-shot learners.
  \item A zero-shot cross-modality transfer capability in CLIP is demonstrated.
  \item A parameter-efficient fine-tuning strategy, BiNor, is proposed to boost CLIP's few-shot visual question answering performance.
\end{itemize}

\begin{figure}[t]
\centering
\includegraphics[width=.99\columnwidth]{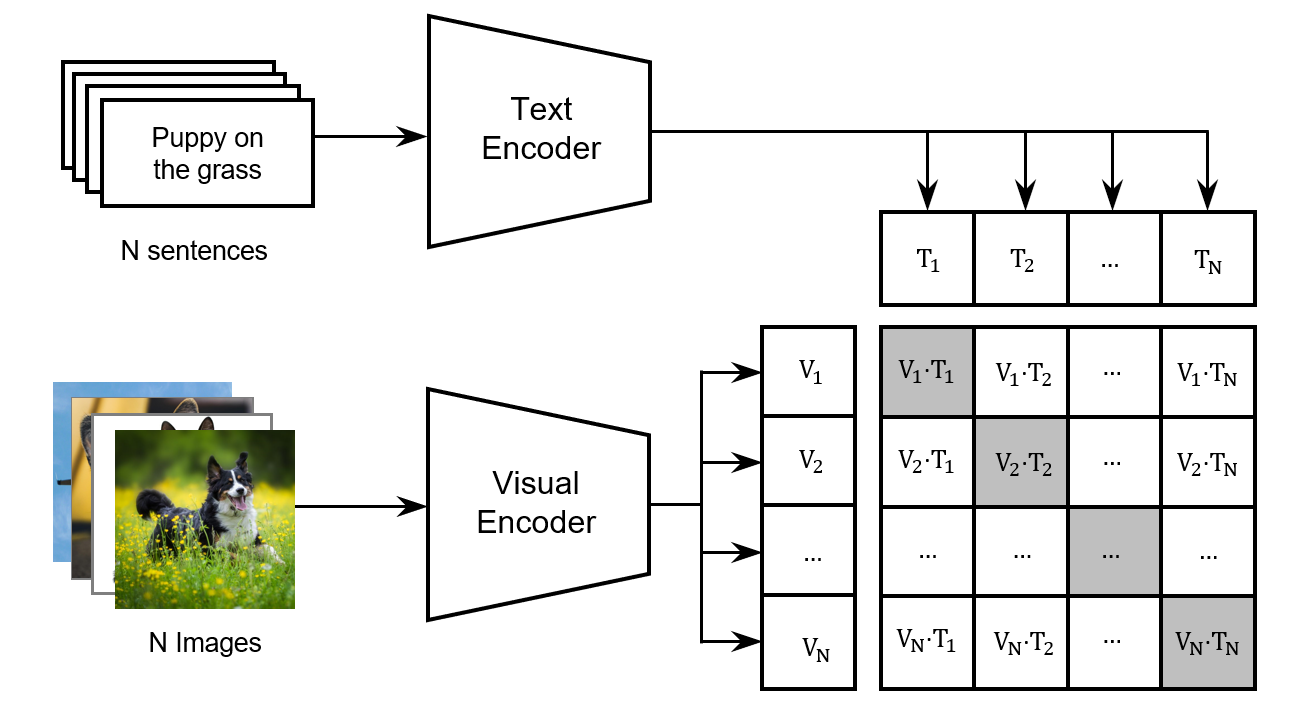}
\caption{ 
CLIP consists of a visual encoder $\mathbb V$, a text encoder $\mathbb T$, and a dot product between their outputs. It is trained to align images and texts with a contrastive loss. The dot product is used as an alignment score.
}
\label{fig:clip}
\end{figure}

\section{Preliminaries}
\label{sec:preliminaries}

\subsection{CLIP}
\label{sec:clip_pre}
CLIP, short for Contrastive Language-Image Pre-training~\cite{radford2021learning}, learns visual representations from natural language supervision. Figure~\ref{fig:clip} shows its key components and the way it works. It consists of a visual encoder $\mathbb V$, e.g. ResNet~\cite{he2016deep} and ViT~\cite{dosovitskiy2020image}, and a text encoder $\mathbb T$, e.g. transformer~\cite{vaswani2017attention}, where they encode images and texts independently. Followed up is a dot-product between the two encoders' outputs, i.e. $\mathbb T(\text{text}) \cdot \mathbb V(\text{image})$, which is used as an alignment score between the input image and text. It is pre-trained to distinguish aligned image-text pairs from randomly combined ones by a contrastive loss. Instead of training on vision benchmarks, CLIP leverages abundant language supervisions from 400 million web-crawled image-text pairs and can conduct a variety of image classification tasks without specific optimizing. However, directly applying CLIP as a vision-language understanding model is still difficult~\cite{kim2021vilt,shen2021much}.

\subsection{Vision-Language Understanding Tasks}
\label{sec:vl_pre}
\paragraph{Visual question answering.} The task of VQA requires the model to answer questions about the details of input images. Following previous work, we experiment on the VQAv2~\cite{goyal2017making} dataset and formulate the task as a classification problem over 3,129 pre-defined most frequent answers. The images in VQAv2 come from Microsoft COCO~\cite{lin2014microsoft}, and there are 65 types of questions in the dataset, such as {\it how many} and {\it what color is}. For answers, there are three types, including {\it yes/no}, {\it number}, and {\it other}.

\begin{figure*}[t]
\centering
\includegraphics[width=\textwidth]{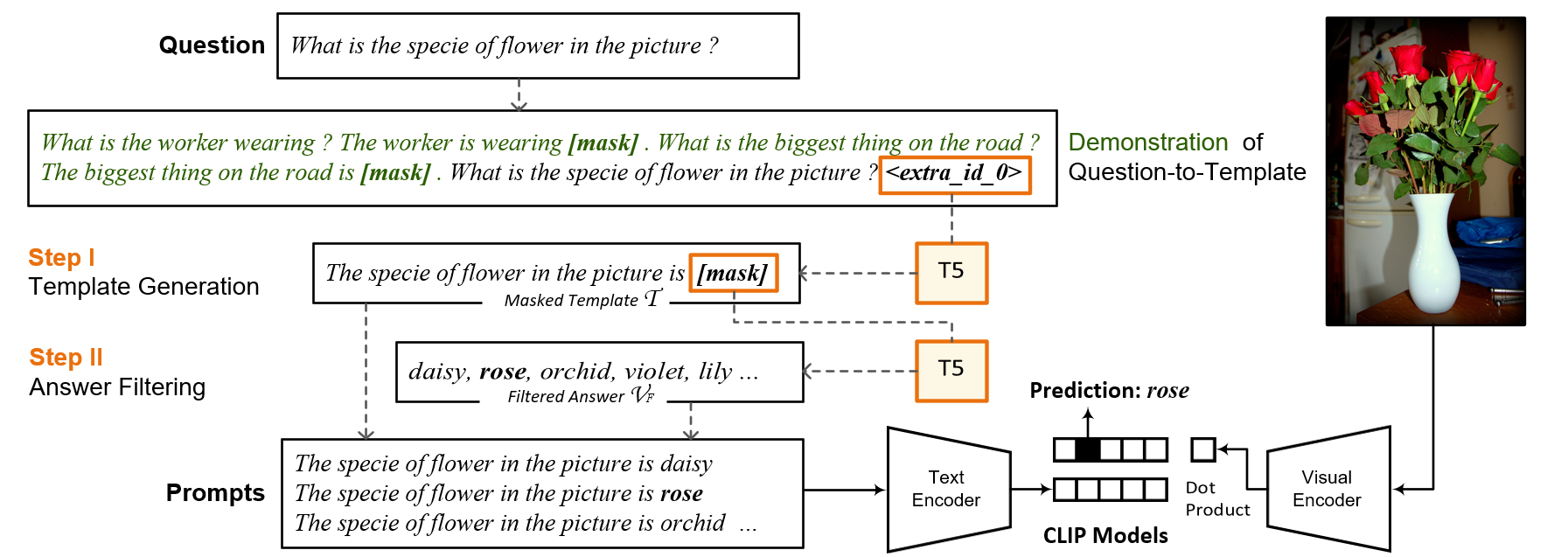}
\caption{ The overall framework of the proposed TAP-C method for zero-shot VQA. TAP-C first generates a masked template from the question by demonstrating examples to T5 and then filters out impossible answers according to the language model. Infilling the masked template with selected answers results in prompts, which could be paired with images to calculate image-text alignment scores by the CLIP. The dashed line denotes the process of prompts generation (\cref{sec:t_vqa}), and the solid line denotes prompting CLIP to conduct zero-shot VQA (\cref{sec:tap-c}).}
\label{fig:2}
\end{figure*}

\paragraph{Visual entailment.} Similar to the natural language inference (NLI), the task of visual entailment predicts the entailment relations, including {\it entailment}, {\it neutral}, and {\it contradiction}, between a premise and a hypothesis. Under the VL setting, the premise in visual entailment is based on the details of an image rather than textual descriptions in NLI. 
The SNLI-VE dataset~\cite{xie2019visual} is adapted from SNLI~\cite{bowman2015large} and replaces SNLI's premises with the images in the Flickr30k dataset~\cite{young2014image}.
Considering the above characteristics, here we leverage the SNLI-VE dataset to verify the zero-shot cross-modality (language and vision) transfer capabilities of the CLIP models.
This zero-shot setting investigates how well the vision and language representations are aligned in CLIP models.



\section{Zero-shot VQA}
\label{sec:zerovqa}
\subsection{A Two-Step Prompt Generation Method}
\label{sec:t_vqa}
Previous works~\cite{kim2021vilt,shen2021much} have found that directly applying CLIP models for zero-shot VL tasks are infeasible. For example, nearly random-chance level zero-shot performances are observed on the VQAv2 dataset by directly applying a ``question: [\textit{question text}] answer: [\textit{answer text}]'' prompt template~\cite{shen2021much}. After rethinking the essence of prompt engineering in CLIP, we can find that the key to a successful zero-shot capability transfer for the VQA task is to mitigate the gap between natural language description and the form of question answering.

Motivated by the above observations, we propose a two-step automatic prompt generation method to enable the zero-shot VQA capabilities in CLIP models, with the assistant of a pre-trained generative T5 model~\cite{raffel2020exploring}. The key ideas of the two-step prompt generation method is illustrated in Figure~\ref{fig:2}: the first step is to convert the question into a masked template $\mathcal T$, and the second step is to filter out impossible answers by language model and get a candidate answer set $\mathcal V_F$. The infilled template connects both the question and answers in a natural description way and thus could be an ideal form of prompt for the VQA task.

\subsubsection*{Step I: Automatic Template Generation}
This step is designed to convert the question into a template, which is a statement with a mask token. 
To tackle the conversion challenge, we explore two ways, including an in-context demonstration method and a dependency parsing based method.

\paragraph{Demonstration to T5.}
The idea of this conversion method is relatively simple: by demonstrating question-to-template (with [mask] token) examples to the language model, the model could implicitly capture the conversion pattern. 
We define a few examples for each question type and convert the questions according to their types. Figure~\ref{fig:2} shows a conversion example. More cases could be found at appendix~\ref{apdx:template-generation}.
Specifically, we use T5~\cite{raffel2020exploring}, a large pre-trained text-to-text Transformer, for the question to template conversion.
T5 is pre-trained to infill the missing spans (replaced by T5 special tokens, e.g. {\it <extra\_id\_0>}) of a sentence. 
We present a concatenation of examples, question, and the {\it <extra\_id\_0>} token to T5 for conditional generation to restore it, and the generated span is our masked template, named as $\mathcal T_\text{demo}$.

\paragraph{Dependency parsing.}
Although the T5 conversion method works well in most situations, it still faces some out-of-coverage problems. To compensate for this shortcoming, we turn to a traditional dependency parsing based way.
This method converts a question to a statement by its part-of-speech tagging and parsing results, where the wh-word, root word, auxiliary, or copula, as well as prepositions and particles that are dependents of the wh-word or the root, are identified, and transformations are performed according to grammar rules. We use the Stanza~\cite{qi2020stanza} to POS tag and parse the question and leave the answer as a mask token. Then the rules\footnote{https://github.com/kelvinguu/qanli} in~\citet{demszky2018transforming} are leveraged to perform the conversion. We name the template obtained in this way as $\mathcal T_\text{parsing}$.

\subsubsection*{Step II: Answer Filtering}
\label{sec:answer_select}
As common sense, ``{\it the specie of a flower}'' can never be a vase. Therefore, leveraging pre-trained language models, which have well learned such concepts during pre-training, to filter out less likely answers would have a positive influence on the final question answering performance. Given a masked template $\mathcal T$, a language model $\mathcal L$, and the answer vocabulary $\mathcal V$, we get the filtered answers $\mathcal V_F$ as:
\begin{equation}
    \mathop{\text{Top-\textit{k}}}_{v \in {\mathcal V}} \left\{ \log P_{\mathcal L} \left(\text{[mask]}=v | {\mathcal T}\right) \right\},
\end{equation}
where the [mask] is the answer span in template $\mathcal T$, and $P_{\mathcal L}$ is the output distribution of the language model. Here we also apply the T5 to infill answers because it makes no assumption about the length and position of the span.
Once we get the template $\mathcal T$ and the filtered answers $\mathcal V_F$, we replace the [mask] token in template $\mathcal T$ with every selected answer in $\mathcal V_F$ to get the prompts $\mathcal P$. 

\subsection{TAP-C Method for VQA}
\label{sec:tap-c}
The proposed method follows a Template-Answer-Prompt then CLIP discrimination pipeline, and thus we name it as \textbf{TAP-C}. To make better use of template $\mathcal T_\text{parsing}$ and $\mathcal T_\text{demo}$, we use an ensemble of both templates by simply setting a threshold for the T5's generation confidence. We prefer to use $\mathcal T_\text{demo}$ but use $\mathcal T_\text{parsing}$ if the generation confidence is low.
Finally, given an image $i$ and the generated prompts $\mathcal P$, the TAP-C method can get a zero-shot VQA prediction by:
\begin{equation}
\label{eq:dot_product}
    \mathop{\max}_{v \in {\mathcal V_F},\ p_v \in \mathcal P} \left\{\mathbb{V} \left( i \right) \cdot \mathbb{T} \left(p_v \right) 
    \right\},
\end{equation}
where $\mathbb{V}$ and $\mathbb{T}$ are the visual and text encoders in CLIP models. The $p_v$ is a prompt generated by the TAP-C method, where the masked template is infilled with answer $v$ from the filtered answer vocabulary $\mathcal V_F$.

\begin{figure}[t]
\centering
\includegraphics[width=.99\columnwidth]{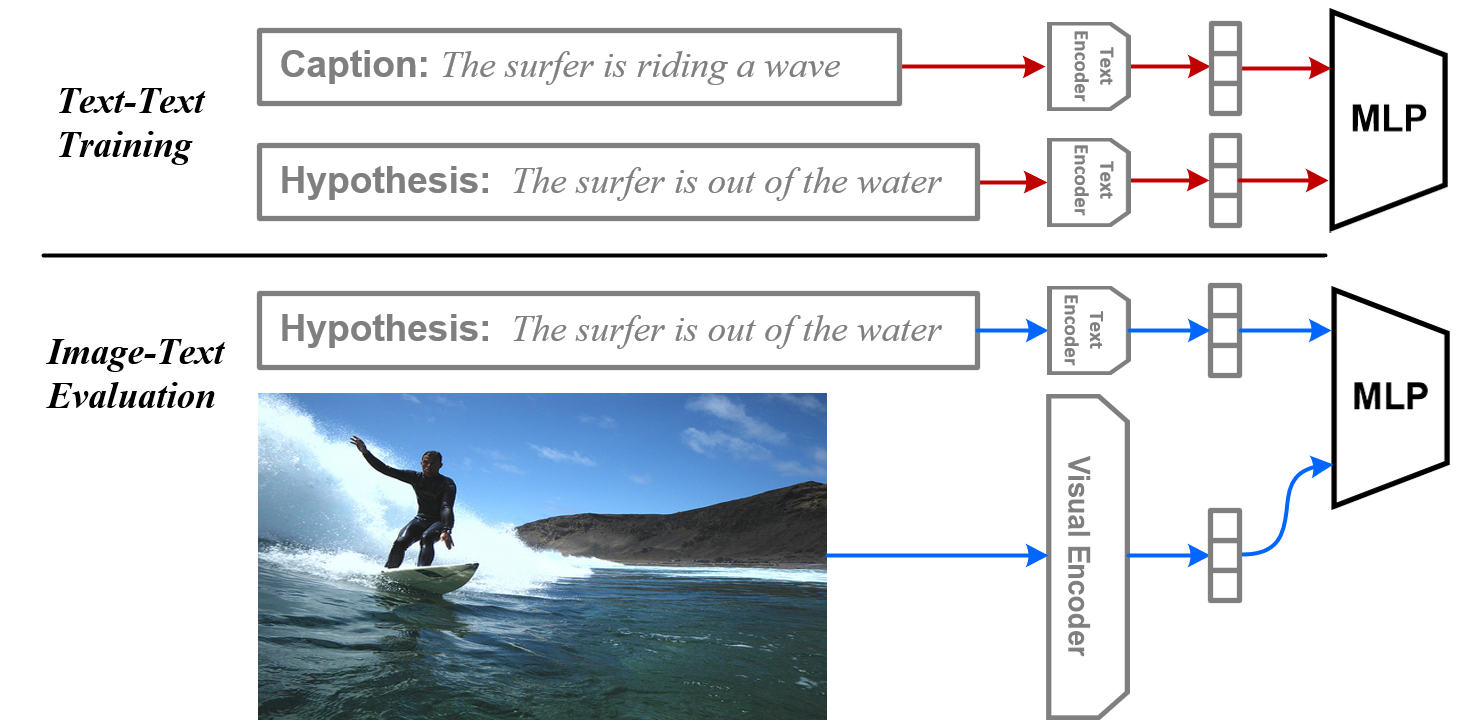}
\caption{Zero-shot cross-modality transfer on visual entailment task. The red line denotes the text-only training process, and the blue line denotes the image-text, i.e., cross-modality, evaluation process. The MLP is shared between training and evaluation, while both encoders in CLIP models are not updated.}
\label{fig:3}
\end{figure}

\section{Zero-shot Cross-modality Transfer}
\label{sec:t_ve}
Recent pre-trained multilingual language models~\cite{wu2019beto,liu2020multilingual,xue2021mt5} have been shown to be successful in transferring representations across different languages. For example, they can be only fine-tuned on a source language and evaluated on various target languages without specific training, yet still achieving good performance. 
On the other hand, the CLIP models achieve strong zero-shot performances on both image-to-text and text-to-image retrieval tasks~\cite{radford2021learning} only through a dot product between vision and language representations, which gives us an intuition that the two modalities are well aligned in the CLIP models. Is there a cross-modality capability between language and vision in the CLIP models, just like the multilingual ones across languages?

To answer the above question, we utilize the visual entailment task (\cref{sec:vl_pre}) to explore the zero-shot cross-modality performance. Figure~\ref{fig:3} briefs the key idea. Specifically, we train an MLP classifier over the fused representations of premise and hypothesis, and the fusion function is:
\begin{equation}
    {\mathrm{fuse}} \left( v_1, v_2 \right) = \left[v_1,v_2,v_1+v_2,v_1-v_2,v_1\cdot v_2\right], \nonumber
\end{equation}
where $v_1$ and $v_2$ are two input vectors. 
During training, text-only premise and hypothesis are used as the input of CLIP text encoder:
\begin{equation}
    {\mathrm{MLP}} \left\{ \mathrm{fuse} \left( {\mathbb T}(\text{pre}_t),\ {\mathbb T}(\text{hyp}_t) \right) \right\},
\end{equation}
where $\mathbb T$ is the CLIP text encoder and is not updated during training. And $\text{pre}_t$ and $\text{hyp}_t$ are the text premise and hypothesis. Then at inference, the premise is in image and is fed into the CLIP visual encoder. The trained MLP is leveraged for prediction:
\begin{equation}
    {\max} \left\{ {\mathrm{MLP}} \left\{ \mathrm{fuse} \left( {\mathbb V}(\text{pre}_i),\ {\mathbb T}(\text{hyp}_t) \right) \right\} \right\},
\end{equation}
where the $\text{pre}_i$ is the image premise and $\mathbb V$ is the CLIP visual encoder. 

\section{Few-shot Learning for VQA}
\label{sec:few}
In this section, We aim to investigate whether the CLIP models could benefit from \textit{few-shot} learning, where we work on the visual question answering task to study it.

\subsection{Setup of Few-shot VQA}
\label{sec:few_def}
Here we briefly define the terminology used in our few-shot visual question answering settings:
\begin{itemize}
    \item \textbf{Number of ways.} Originally, it is defined as the distinct classes in a task. However, rather than defining a 3,129-way task according to the answer vocabulary, we define the number of ways as question type times answer type (\cref{sec:vl_pre}), i.e., 65$ \times$3=195 ways, to ensure the model's generalization ability where it can answer a type of questions.
    \item \textbf{Number of shots.} The number of distinct examples in each way. Here a shot is an image along with the question and the answer.
    \item \textbf{Support set and query set.} Before training, we will sample a 195-way K-shot subset from the VQAv2 training set, and thus there are 195$\times$K distinct examples available during few-shot learning. In each training epoch, we select $C$ ways out of 195 ways for parameter optimizing and divide \textit{k} shots in each way into support set and query set with a fixed proportion. The support set is used for model training, and the query set is used for performance evaluation (similar to a typical dev set).
\end{itemize}

\subsection{Parameter-efficient Fine-tuning}
\label{sec:tn_stra}
Under the few-shot setting, our goal is to make the CLIP models learn from N-way K-shot examples and improve the zero-shot VQA performance.
Specifically, we identify only a very small set of parameters in CLIP models (about 0.3 million out of over 100 million, details in appendix~\ref{apdx:number-para}), including the \textit{bias} term and \textit{normalization} term, to be optimized. For either the BatchNorm in ResNet or the LayerNorm in Transformer, the normalization could be uniformly denoted as:
\begin{equation}
    y = \frac{x-\text{E}(x)}{\sqrt{\text{Var}(x)+\epsilon}}\cdot \gamma + \beta,
\end{equation}
where $x$ and $y$ are the mini-batched input and output, and the $\gamma$ and $\beta$ are learned parameters. And for all the linear layers and projection layers in CLIP models, they could be denoted as:
\begin{equation}
    o = w \cdot h + bias,
\end{equation}
where $h$ and $o$ are the input and output vectors. We define the learnable parameter set as:
\begin{equation}
\label{eq:binor}
    {\text P}_{learn} =  \{ bias , \gamma , \beta \}.
\end{equation}
We optimize the \textbf{Bi}as and \textbf{Nor}malization (BiNor) parameters on the few-shot examples with a standard cross-entropy loss over the dot products from each image-prompt pair (Eq.\ref{eq:dot_product}).

Besides, when there are a few examples available, we could also leverage an in-context demonstration manner to improve the performance of the \textit{answer filtering} process in TAP-C (\cref{sec:answer_select}) by:
\begin{equation}
\label{eq:demo_answer}
    \mathop{\text{Top-\textit{k}}}_{v \in {\mathcal V}} \left\{ \log P_{\mathcal L} \left(\text{[mask]}=v\ |\ {[\mathcal D ,\mathcal T}]\right) \right\},
\end{equation}
where the $\mathcal D$ denotes the demonstrations. $\mathcal D$ is similar to template $\mathcal T$ but has been infilled with the answers, and it is sampled from the same type of question in the available few-shot examples. The resulting filtered vocabulary is noted as $\mathcal V_\text{demo}$.
We report the few-shot training procedure in appendix~\ref{apdx:algorithm}.

\section{Experiments}
\label{sec:experiments}

\subsection{Experimental Settings}
\label{sec:exp_setup}

\paragraph{Datasets.} For visual question answering and visual entailment, we carry out experiments on the VQAv2~\cite{goyal2017making} and the SNLI-VE~\cite{xie2019visual} datasets, respectively. We report the statistics of the two datasets in appendix~\ref{apdx:dataset}. For the VQA task's evaluation, we follow the \textit{Frozen} model~\cite{tsimpoukelli2021multimodal} to calculate the \textit{vqa scores} on the VQAv2 validation set. For visual entailment, we calculate the \textit{accuracy} on both validation and test sets through the \textit{sklearn} toolkit.

\paragraph{CLIP models.} According to the types of visual encoders, e.g. ResNet or ViT, CLIP models have different variants, resulting in a significant difference in the number of learnable \textit{bias} and \textit{normalization} parameters. We report the number of learnable parameters of CLIP variants in appendix~\ref{apdx:number-para}. We select two best performing (and publicly available) variants from two kinds of visual encoders, including the CLIP Res50x16 and the CLIP ViT-B/16, to empirically study their zero-shot and few-shot vision-language understanding performances by applying our transferring methods (\cref{sec:zerovqa,sec:t_ve,sec:few}).

\subsection{Results of Zero-shot VQA}
\label{sec:exp_zerovqa}

As previous VL models heavily rely on object detection sub-modules, it is not feasible to directly apply them under the zero-shot setting. Here we setup zero-shot VL baselines from two latest works:
\begin{itemize}
    \item \textbf{Frozen.} \textit{Frozen}~\cite{tsimpoukelli2021multimodal} prompts a seven-billion-parameter 32-layer language model with image representations. It is trained on aligned image-caption data and is also the first model that shows promising zero-shot and few-shot VQA performances.
    \item \textbf{Question irrelevant prompt.} \citet{shen2021much} explored directly prompting the CLIP models for the VQA task. They used a ``question: [\textit{question text}] answer: [\textit{answer text}]'' template, together with the prompt engineering of image classification, to prepare prompts. The resulting prompts are irrelevant to questions, and thus we note this method as QIP. 
\end{itemize}

We report the zero-shot VQA results in Table~\ref{tab:zero-vqa}.
The experimental results verify our hypothesis (\cref{sec:t_vqa}) that the prompts of CLIP should be used to describe the labels rather than the tasks. As we can see, the question irrelevant prompting methods simply present the task description and answers to the CLIP models and only get barely better than random guess results. In contrast, by converting questions into templates and filtering answers with pre-trained language models, our TAP-C method enables CLIP models a strong zero-shot capability on the VQA task, even compared with the seven-billion-parameter \textit{Frozen} zero-shot model.

\begin{table}[t]
\centering
\resizebox{.99\columnwidth}{!}{%
\begin{tabular}{@{}l|rrrr@{}}
\toprule
Zero-shot Methods & \multicolumn{1}{c}{Yes/No} & \multicolumn{1}{c}{Number} & \multicolumn{1}{c}{Other} & \multicolumn{1}{c}{All} \\ \midrule
\textbf{Frozen}~\cite{tsimpoukelli2021multimodal}   & -    & -    & -   & 29.50 \\ 
\midrule
\textbf{QIP}~\cite{shen2021much}  &        &       &      &       \\
~~w/ CLIP$_\text{Res101}$                    & 53.01  & 6.67  & 0.96 & 21.26 \\
~~w/ CLIP$_\text{Res50x16}$                  & 56.16  & 9.76  & 1.39 & 23.07 \\ 
~~w/ CLIP$_\text{ViT-B/16}$                  & 53.89	 & 7.67  & 0.70 & 21.40 \\
\midrule
\textbf{TAP-C (Ours)}          &        &       &      &       \\
~~w/ CLIP$_\text{Res50x16}$                  & \textbf{71.65} & 18.74 & 18.22 & 38.36 \\
~~w/ CLIP$_\text{ViT-B/16}$                  & 71.38 & \textbf{20.95} & \textbf{18.55} & \textbf{38.72} \\ \bottomrule
\end{tabular}%
}
\caption{Zero-shot results on the VQAv2 validation set.}
\label{tab:zero-vqa}
\end{table}

\begin{table}[t]
\centering
\resizebox{\columnwidth}{!}{%
\begin{tabular}{@{}l|cc|c@{}}
\toprule
\textbf{Training} & \multicolumn{2}{c|}{Text + Text} & Image + Text \\ \midrule
\textbf{Evaluation} & Image + Text & Image Masked & Text + Text \\ \midrule
Majority               & 33.37 / 33.37 & 33.37 / 33.37 & 33.37 / 33.37 \\
CLIP$_\text{ViT-B/16}$ & 64.11 / 64.66 & 35.05 / 35.69 & 65.97 / 66.23 \\
CLIP$_\text{Res101}$ & 64.29 / 64.86 & 36.27 / 35.34 & 65.67 / 66.28 \\
CLIP$_\text{Res50x16}$ & 67.24 / 66.63 & 36.36 / 36.05 & 67.64 / 68.18 \\ \bottomrule
\end{tabular}%
}
\caption{Zero-shot cross-modality transfer results on the SNLI-VE valid and test set (\textit{valid acc} / \textit{test acc}).}
\label{tab:result_ve}
\end{table}

\begin{table*}[t]
\centering
\resizebox{\textwidth}{!}{%
\begin{tabular}{@{}ll|rrrrrrrr@{}}
\toprule
\multicolumn{2}{l|}{\textbf{Fully Supervised Results on Test-Dev}} & \multicolumn{1}{l}{} & \multicolumn{1}{l}{} & Y/N$_{\text{full}}$ & NUM$_{\text{full}}$ & OTHER$_{\text{full}}$ & ALL$_{\text{full}}$ & \multicolumn{1}{l}{} & \multicolumn{1}{l}{} \\ \midrule
\multicolumn{2}{l|}{Mcan~\cite{yu2019deep}} & \multicolumn{1}{l}{} & \multicolumn{1}{l}{} & 86.82 & 53.26 & 60.72 & 70.63 & \multicolumn{1}{l}{} & \multicolumn{1}{l}{} \\
\multicolumn{2}{l|}{CLIP-ViL$_{\text{p}}$~\cite{shen2021much}} & \multicolumn{1}{l}{} & \multicolumn{1}{l}{} & - & - & - & 76.48 & \multicolumn{1}{l}{} & \multicolumn{1}{l}{} \\ \midrule
\multicolumn{2}{l|}{\textbf{Few-shot Results on Validation Set}} & Y/N$_{\text{K=1}}$ & NUM$_{\text{K=1}}$ & OTHER$_{\text{K=1}}$ & \multicolumn{1}{r|}{ALL$_{\text{K=1}}$} & Y/N$_{\text{K=4}}$ & NUM$_{\text{K=4}}$ & OTHER$_{\text{K=4}}$ & ALL$_{\text{K=4}}$ \\ \midrule
\multicolumn{2}{l|}{Frozen$_\text{blind}$~\cite{tsimpoukelli2021multimodal}} & - & - & - & \multicolumn{1}{r|}{33.50} & - & - & - & 33.30 \\
\multicolumn{2}{l|}{Frozen~\cite{tsimpoukelli2021multimodal}} & - & - & - & \multicolumn{1}{r|}{35.70} & - & - & - & 38.20 \\
\multicolumn{2}{l|}{\textbf{TAP-C$_\text{ViT-B/16}$} \textbf{(Ours)}} & 71.03 & 29.72 & 25.73 & \multicolumn{1}{r|}{\textbf{43.27}} & 71.53 & 31.40 & 28.36 & 44.98 \\
\multicolumn{2}{l|}{~~w/o $\mathcal V_\text{demo}$} & 71.03 & \textbf{29.74} & 19.01 & \multicolumn{1}{r|}{39.96} & 71.53 & \textbf{31.45} & 21.78 & 41.74 \\
\multicolumn{2}{l|}{\textbf{TAP-C$_\text{Res50x16}$} \textbf{(Ours)}} & \textbf{71.77} & 26.75 & \textbf{25.88} & \multicolumn{1}{r|}{43.24} & \textbf{71.86} & 27.86 & \textbf{30.86} & \textbf{45.87} \\
\multicolumn{2}{l|}{~~w/o $\mathcal V_\text{demo}$} & \textbf{71.77} & 26.73 & 19.97 & \multicolumn{1}{r|}{40.32} & \textbf{71.86} & 27.92 & 22.43 & 41.72 \\ \midrule
\multicolumn{2}{l|}{} & Y/N$_{\text{K=16}}$ & NUM$_{\text{K=16}}$ & OTHER$_{\text{K=16}}$ & \multicolumn{1}{r|}{ALL$_{\text{K=16}}$} & Y/N$_{\text{K=32}}$ & NUM$_{\text{K=32}}$ & OTHER$_{\text{K=32}}$ & ALL$_{\text{K=32}}$ \\ \midrule
\multicolumn{2}{l|}{\textbf{TAP-C$_\text{ViT-B/16}$} \textbf{(Ours)}} & \textbf{73.05} & \textbf{31.46} & 32.13 & \multicolumn{1}{r|}{47.42} & \textbf{73.60} & \textbf{32.55} & 35.02 & 49.19 \\
\multicolumn{2}{l|}{~~w/o $\mathcal V_\text{demo}$} & \textbf{73.05} & 31.44 & 25.08 & \multicolumn{1}{r|}{43.94} & \textbf{73.60} & 32.52 & 26.95 & 45.21 \\
\multicolumn{2}{l|}{\textbf{TAP-C$_\text{Res50x16}$} \textbf{(Ours)}} & 72.98 & 29.96 & \textbf{35.58} & \multicolumn{1}{r|}{\textbf{48.89}} & 73.51 & 31.56 & \textbf{37.35} & \textbf{50.18} \\
\multicolumn{2}{l|}{~~w/o $\mathcal V_\text{demo}$} & 72.98 & 29.87 & 26.53 & \multicolumn{1}{r|}{44.42} & 73.51 & 31.70 & 28.26 & 45.71 \\ \bottomrule
\end{tabular}%
}
\caption{Few-shot VQA results under different \textit{k} values, along with two fully supervised models' performance as references. The $\mathcal V_\text{demo}$ enhances answer filtering by few-shot demonstration to T5, details in Eq.\ref{eq:demo_answer}. Our few-shot method not only outperform \textit{Frozen}, but also achieves stable improvements under different \textit{k} values.}
\label{tab:few_vqa}
\end{table*}

\subsection{Zero-shot Cross-modality Transfer}
\label{sec:exp_zero_ve}
We report the zero-shot cross-modality transfer results in Table~\ref{tab:result_ve}. We first investigate the language to vision transfer capability. As introduced in \cref{sec:t_ve}, we train a classifier on the text-only SNLI-VE dataset where the image is replaced by its caption. At inference, the trained classifier is evaluated by taking the image and text as inputs. As shown in the first group of results, after solely trained on text-text (caption as the premise) entailment data, different CLIP variants could successfully gain a similar discriminative ability under the image-text setting. To ensure that the above results are indeed transferring from language to vision, we made a double check by masking out the images at inference time, and the results are reported at \textit{Image Masked}. As we can see, the results are similar to a random guess of three relations, indicating the images are of importance in the cross-modality evaluation.

Now that we have observed the language to vision transferring capability in CLIP models, we further investigate whether there is also a vision to language transfer capability. We conduct a similar experiment but train the classifier on the original SNLI-VE dataset, i.e., image premise and text hypothesis. At inference, we evaluate the classifier with the text-only valid and test data. The results are reported in Table~\ref{tab:result_ve}, which confirms the vision to language capability. Since text data are usually much cheaper than visual data, the first kind of transferring tends to be more promising in practice.

\subsection{Results of Few-shot VQA}
\label{sec:exp_few}
We report the few-shot VQA results in Table~\ref{tab:few_vqa}. We take the \textit{Frozen} model and the image blacked out \textit{Frozen}$_{\text{blind}}$ as baselines. Under different \textit{k}, our methods could always learn from limited training examples and improve over the zero-shot results, which confirms that CLIP models could be VL few-shot learners.
With the increase of the number of shots, significant performance gains are observed in \textit{other} category, which concurs with our intuition: as we sample examples from each question type, most answers in \textit{other} category are not revealed to the model. As a result, the model could always learn to improve.
Similarly, presenting examples to the T5 could also improve the answer filtering process, leading to significant performance gains over the \textit{other} category.
In contrast, the score of \textit{number} category improves significantly when the model just begins to see some training examples while slowing down as \textit{k} continues to increase.

\begin{table}[t]
\centering
\resizebox{.99\columnwidth}{!}{%
\begin{tabular}{@{}l|llll@{}}
\toprule
Zero-shot Methods & \multicolumn{1}{c}{Yes/No} & \multicolumn{1}{c}{Number} & \multicolumn{1}{c}{Other} & \multicolumn{1}{c}{All} \\ \midrule

\textbf{TAP-C$_\text{ViT-B/16}$}              & 71.38 & 20.95 & 18.55 & 38.72 \\
~~w/o $\mathcal T_\text{demo}$     & 71.36 & 20.86 & 17.96$^\dagger$ & 38.41$^\dagger$ \\
~~w/o $\mathcal T_\text{parsing}$     & 70.82$^\dagger$ & 19.86$^\dagger$ & 18.40 & 38.29$^\dagger$ \\ \bottomrule
\end{tabular}%
}
\caption{Ablation results of template generation methods. $\mathcal T_\text{demo}$ and $\mathcal T_\text{parsing}$ denote the T5 demonstration template and dependency parsing template. ``$\dagger$'' means statistically significant difference (2-tailed t-test, p<0.01).
}
\label{tab:template-ablation}
\end{table}

\begin{table}[t]
\centering
\resizebox{\columnwidth}{!}{%
\begin{tabular}{@{}l|lll@{}}
\toprule
Methods & K=0 & K=4 & K=32 \\ \midrule
\textbf{TAP-C$_\text{ViT-B/16}$} & 38.72 & 44.98 & 49.19 \\
~~w/o a.filt.          & 32.57 (16\%) & 35.07 (22\%) & 40.21 (18\%) \\
~~w/o t.gen. + a.filt.   & 21.40 (45\%) & 22.59 (50\%) & 23.76 (52\%) \\ \midrule
\textbf{TAP-C$_\text{Res50x16}$} & 38.36 & 45.87 & 50.18 \\
~~w/o a.filt. & 32.43 (16\%) & 34.56 (25\%) & 40.97 (18\%) \\
~~w/o t.gen. + a.filt. & 23.07 (40\%) & 23.98 (48\%) & 24.86 (51\%) \\ \bottomrule
\end{tabular}%
}
\caption{Ablation results of TAP-C. In brackets is the percentage of VQA performance degradation. When both steps are all removed, the few-shot learning is performed on CLIP with question irrelevant prompts.}
\label{tab:ablation}
\end{table}

\begin{table*}[t]
\centering
\resizebox{\textwidth}{!}{%
\begin{tabular}{@{}l|ccc|ccc|ccc|ccc@{}}
\toprule
\multirow{2}{*}{Few-shot Mehtods} & \multicolumn{3}{c|}{K=1} & \multicolumn{3}{c|}{K=4} & \multicolumn{3}{c|}{K=16} & \multicolumn{3}{c}{K=32} \\ \cmidrule(l){2-13} 
 & Full-FT & BitFit & BiNor & Full-FT & BitFit & BiNor & Full-FT & BitFit & BiNor & Full-FT & BitFit & BiNor \\ \midrule
TAP-C$_\text{ViT-B/16}$   & 37.78$^\ast$ & 42.96 & \textbf{43.27} & 38.30$^\ast$ & 44.77 & \textbf{44.98} & 39.99 & 46.80 & \textbf{47.42} & 40.35 & 47.78 & \textbf{49.19} \\
TAP-C$_\text{Res101}$     & 36.63$^\ast$ & 42.21 & \textbf{42.98} & 37.92$^\ast$ & 43.02 & \textbf{44.72} & 39.42 & 44.83 & \textbf{47.43} & 39.58 & 45.59 & \textbf{48.86} \\
TAP-C$_\text{Res50x16}$   & 35.96$^\ast$ & \textbf{43.38} & 43.24 & 38.03$^\ast$ & 44.30 & \textbf{45.87} & 39.84 & 46.57 & \textbf{48.89} & 40.42 & 47.74 & \textbf{50.18} \\ \bottomrule
\end{tabular}%
}
\caption{The performance comparisons of different fine-tuning strategies on the few-shot VQA task. Results marked with ``$\ast$'' are lower than the zero-shot performance. Full-FT is short for full fine-tuning. Besides BiNor's good performance, it improves ResNet CLIPs more significantly due to the number of normalization parameters.}
\label{tab:analysis-bitfit}
\end{table*}

\subsection{Analyses and Discussion}
\label{sec:exp_analysis}
\paragraph{The effects of template generation methods.} Our TAP-C method uses an ensemble of dependency parsing template $\mathcal T_\text{parsing}$ and T5 demonstration template $\mathcal T_\text{demo}$. Here we investigate whether it is necessary to use such an ensemble. We report the ablation results of two templates in Table~\ref{tab:template-ablation}. The results show that the two templates have different effects over different questions, and the ensemble could make the best use of their advantages.

\paragraph{The effects of two steps in TAP-C.} The TAP-C method generates prompts through template generation (\textbf{t.gen.}) and answer filtering (\textbf{a.filt.}). Here we quantify how much each step contributes to the final zero/few-shot VQA performances.
We report the ablation results in Table~\ref{tab:ablation}. When we remove the \textit{answer filtering} step (w/o a.filt.), both the zero-shot and few-shot performances generally fall by about 20\%, but the models still retain some few-shot learning capabilities. 
We further remove the \textit{template generation} step and only use question irrelevant templates: all results are nearly cut in half, indicating the importance of considering questions in both zero-shot and few-shot scenarios.

\paragraph{Comparisons of fine-tuning methods.} We only update the bias and normalization parameters during few-shot learning (\cref{sec:tn_stra}). To investigate whether our BiNor fine-tuning strategy works well, we compare BiNor with two fine-tuning methods:
1) \textbf{Full-FT} (Full fine-tuning), which updates all parameters in the model.
2) \textbf{BitFit}~\cite{ben2021bitfit}, which only updates the bias-terms in all model layers. 
We report the comparison results in Table~\ref{tab:analysis-bitfit}. Both BiNor and BitFit significantly outperform the full fine-tuning way: millions of parameters are very easy to overfit to a few training examples. When \textit{k} is small, the performance differences between BiNor and BitFit are very small. When \textit{k} becomes larger, BiNor begins to outperform BitFit with a noticeable margin.
Our BiNor fine-tuning strategy is similar to the BitFit but differs in that it also updates the normalization parameters, which would grant the ResNet CLIP models better flexibility to adapt to new examples due to their larger number of batch normalization parameters.
For the specific number of different parameters in each CLIP variant, please refer to the appendix~\ref{apdx:number-para}.

\paragraph{Limitations of TAP-C.}
The proposed TAP-C method explores CLIP models' potential to conduct zero/few-shot VQA tasks. However, we also found several limitations that hinder further improving the few-shot performance, which could be rooted in the CLIP models. First, CLIP models struggle with counting the number of fine-grained objects in an image, especially counting from a small area of the image. This shortcoming can hardly be improved by any kind of language knowledge. Besides, the CLIP models perform poorly in distinguishing subtle semantic differences. For example,  when asked ``what is the man in the background doing?'', all the experimented CLIP models give predictions of the man ``in the foreground''. Under such cases, even if the TAP-C method perfectly converts the question into a prompt, the final results would still be wrong. Nevertheless, We believe this issue could be well addressed by enhancing CLIP models with a stronger text encoder, and we will make explorations in future work.

\section{Related Work}

\paragraph{Vision-language few-shot learning.} Leveraging aligned caption data, vision-language models pre-trained by an image-text discriminative loss have recently enabled strong zero-shot generalization on image classification and cross-modality retrieval tasks~\cite{jia2021scaling,radford2021learning}. Different from the discriminative manner,~\citet{tsimpoukelli2021multimodal} prompt a large frozen language model with vision prefix in a generative way, which is the first vision-language few-shot model.

\paragraph{Language model prompting.} This work is also inspired by the line of research in language model prompting~\cite{liu2021pre}. Initialized by the GPT series~\cite{radford2018improving,radford2019language,brown2020language}, prompting has become a popular manner to mining knowledge from pre-trained language models~\cite{petroni2019language} in a zero-shot or few-shot way~\cite{shin2020eliciting,gao-etal-2021-making,qin2021learning}. Besides mining knowledge from the language model, PET work~\cite{schick2021exploiting,schick2021s} presents a semi-supervised prompting method for improving few-shot language understanding performance. 

\section{Conclusions}
In this work, we empirically studied how to transfer CLIP models into vision-language understanding tasks. We first explored the CLIP models' zero-shot VQA capability by leveraging language prompts and further proposed a parameter-efficient fine-tuning method to boost the few-shot performance. We also demonstrate a zero-shot cross-modality transfer capability of CLIP models on the visual entailment task. Experiments and analyses on VQAv2 and SNLI-VE confirm that the CLIP models can be good VL few-shot learners.

\section*{Acknowledgements}
Haoyu Song, Wei-Nan Zhang, and Ting Liu are supported by the Science and Technology Innovation 2030 Major Project of China (No.2020AAA0108605), National Natural Science Foundation of China (No.62076081, No.61772153, and No.61936010), and Natural Science Foundation of Heilongjiang (No.YQ2021F006).

\bibliography{clipvl}

\begin{thebibliography}{41}
\expandafter\ifx\csname natexlab\endcsname\relax\def\natexlab#1{#1}\fi

\bibitem[{Antol et~al.(2015)Antol, Agrawal, Lu, Mitchell, Batra, Zitnick, and
  Parikh}]{antol2015vqa}
Stanislaw Antol, Aishwarya Agrawal, Jiasen Lu, Margaret Mitchell, Dhruv Batra,
  C~Lawrence Zitnick, and Devi Parikh. 2015.
\newblock Vqa: Visual question answering.
\newblock In \emph{Proceedings of the IEEE international conference on computer
  vision}, pages 2425--2433.

\bibitem[{Ben~Zaken et~al.(2021)Ben~Zaken, Ravfogel, and
  Goldberg}]{ben2021bitfit}
Elad Ben~Zaken, Shauli Ravfogel, and Yoav Goldberg. 2021.
\newblock Bitfit: Simple parameter-efficient fine-tuning for transformer-based
  masked language-models.
\newblock \emph{arXiv e-prints}, pages arXiv--2106.

\bibitem[{Bowman et~al.(2015)Bowman, Angeli, Potts, and
  Manning}]{bowman2015large}
Samuel Bowman, Gabor Angeli, Christopher Potts, and Christopher~D Manning.
  2015.
\newblock A large annotated corpus for learning natural language inference.
\newblock In \emph{Proceedings of the 2015 Conference on Empirical Methods in
  Natural Language Processing}, pages 632--642.

\bibitem[{Brown et~al.(2020)Brown, Mann, Ryder, Subbiah, Kaplan, Dhariwal,
  Neelakantan, Shyam, Sastry, Askell et~al.}]{brown2020language}
Tom~B Brown, Benjamin Mann, Nick Ryder, Melanie Subbiah, Jared Kaplan, Prafulla
  Dhariwal, Arvind Neelakantan, Pranav Shyam, Girish Sastry, Amanda Askell,
  et~al. 2020.
\newblock Language models are few-shot learners.
\newblock \emph{arXiv preprint arXiv:2005.14165}.

\bibitem[{Chen et~al.(2020)Chen, Li, Yu, El~Kholy, Ahmed, Gan, Cheng, and
  Liu}]{chen2020uniter}
Yen-Chun Chen, Linjie Li, Licheng Yu, Ahmed El~Kholy, Faisal Ahmed, Zhe Gan,
  Yu~Cheng, and Jingjing Liu. 2020.
\newblock Uniter: Universal image-text representation learning.
\newblock In \emph{European conference on computer vision}, pages 104--120.
  Springer.

\bibitem[{Demszky et~al.(2018)Demszky, Guu, and
  Liang}]{demszky2018transforming}
Dorottya Demszky, Kelvin Guu, and Percy Liang. 2018.
\newblock Transforming question answering datasets into natural language
  inference datasets.
\newblock \emph{arXiv preprint arXiv:1809.02922}.

\bibitem[{Deng et~al.(2009)Deng, Dong, Socher, Li, Li, and
  Fei-Fei}]{deng2009imagenet}
Jia Deng, Wei Dong, Richard Socher, Li-Jia Li, Kai Li, and Li~Fei-Fei. 2009.
\newblock Imagenet: A large-scale hierarchical image database.
\newblock In \emph{2009 IEEE conference on computer vision and pattern
  recognition}, pages 248--255. Ieee.

\bibitem[{Devlin et~al.(2019)Devlin, Chang, Lee, and
  Toutanova}]{devlin2019bert}
Jacob Devlin, Ming-Wei Chang, Kenton Lee, and Kristina Toutanova. 2019.
\newblock Bert: Pre-training of deep bidirectional transformers for language
  understanding.
\newblock In \emph{Proceedings of the 2019 Conference of the North American
  Chapter of the Association for Computational Linguistics: Human Language
  Technologies, Volume 1 (Long and Short Papers)}, pages 4171--4186.

\bibitem[{Dosovitskiy et~al.(2020)Dosovitskiy, Beyer, Kolesnikov, Weissenborn,
  Zhai, Unterthiner, Dehghani, Minderer, Heigold, Gelly
  et~al.}]{dosovitskiy2020image}
Alexey Dosovitskiy, Lucas Beyer, Alexander Kolesnikov, Dirk Weissenborn,
  Xiaohua Zhai, Thomas Unterthiner, Mostafa Dehghani, Matthias Minderer, Georg
  Heigold, Sylvain Gelly, et~al. 2020.
\newblock An image is worth 16x16 words: Transformers for image recognition at
  scale.
\newblock In \emph{International Conference on Learning Representations}.

\bibitem[{Gao et~al.(2021)Gao, Fisch, and Chen}]{gao-etal-2021-making}
Tianyu Gao, Adam Fisch, and Danqi Chen. 2021.
\newblock Making pre-trained language models better few-shot learners.
\newblock In \emph{Proceedings of the ACL 2021 (Volume 1: Long Papers)}, pages
  3816--3830, Online. Association for Computational Linguistics.

\bibitem[{Goyal et~al.(2017)Goyal, Khot, Summers-Stay, Batra, and
  Parikh}]{goyal2017making}
Yash Goyal, Tejas Khot, Douglas Summers-Stay, Dhruv Batra, and Devi Parikh.
  2017.
\newblock Making the v in vqa matter: Elevating the role of image understanding
  in visual question answering.
\newblock In \emph{Proceedings of the IEEE Conference on Computer Vision and
  Pattern Recognition}, pages 6904--6913.

\bibitem[{He et~al.(2016)He, Zhang, Ren, and Sun}]{he2016deep}
Kaiming He, Xiangyu Zhang, Shaoqing Ren, and Jian Sun. 2016.
\newblock Deep residual learning for image recognition.
\newblock In \emph{Proceedings of the IEEE conference on computer vision and
  pattern recognition}, pages 770--778.

\bibitem[{Jia et~al.(2021)Jia, Yang, Xia, Chen, Parekh, Pham, Le, Sung, Li, and
  Duerig}]{jia2021scaling}
Chao Jia, Yinfei Yang, Ye~Xia, Yi-Ting Chen, Zarana Parekh, Hieu Pham, Quoc~V
  Le, Yunhsuan Sung, Zhen Li, and Tom Duerig. 2021.
\newblock Scaling up visual and vision-language representation learning with
  noisy text supervision.
\newblock \emph{arXiv preprint arXiv:2102.05918}.

\bibitem[{Kim et~al.(2021)Kim, Son, and Kim}]{kim2021vilt}
Wonjae Kim, Bokyung Son, and Ildoo Kim. 2021.
\newblock Vilt: Vision-and-language transformer without convolution or region
  supervision.
\newblock \emph{arXiv preprint arXiv:2102.03334}.

\bibitem[{Kuznetsova et~al.(2020)Kuznetsova, Rom, Alldrin, Uijlings, Krasin,
  Pont-Tuset, Kamali, Popov, Malloci, Kolesnikov et~al.}]{kuznetsova2020open}
Alina Kuznetsova, Hassan Rom, Neil Alldrin, Jasper Uijlings, Ivan Krasin, Jordi
  Pont-Tuset, Shahab Kamali, Stefan Popov, Matteo Malloci, Alexander
  Kolesnikov, et~al. 2020.
\newblock The open images dataset v4.
\newblock \emph{International Journal of Computer Vision}, 128(7):1956--1981.

\bibitem[{Lin et~al.(2014)Lin, Maire, Belongie, Hays, Perona, Ramanan,
  Doll{\'a}r, and Zitnick}]{lin2014microsoft}
Tsung-Yi Lin, Michael Maire, Serge Belongie, James Hays, Pietro Perona, Deva
  Ramanan, Piotr Doll{\'a}r, and C~Lawrence Zitnick. 2014.
\newblock Microsoft coco: Common objects in context.
\newblock In \emph{European conference on computer vision}, pages 740--755.
  Springer.

\bibitem[{Liu et~al.(2021)Liu, Yuan, Fu, Jiang, Hayashi, and
  Neubig}]{liu2021pre}
Pengfei Liu, Weizhe Yuan, Jinlan Fu, Zhengbao Jiang, Hiroaki Hayashi, and
  Graham Neubig. 2021.
\newblock Pre-train, prompt, and predict: A systematic survey of prompting
  methods in natural language processing.
\newblock \emph{arXiv preprint arXiv:2107.13586}.

\bibitem[{Liu et~al.(2020)Liu, Gu, Goyal, Li, Edunov, Ghazvininejad, Lewis, and
  Zettlemoyer}]{liu2020multilingual}
Yinhan Liu, Jiatao Gu, Naman Goyal, Xian Li, Sergey Edunov, Marjan
  Ghazvininejad, Mike Lewis, and Luke Zettlemoyer. 2020.
\newblock Multilingual denoising pre-training for neural machine translation.
\newblock \emph{Transactions of the Association for Computational Linguistics},
  8:726--742.

\bibitem[{Lu et~al.(2019)Lu, Batra, Parikh, and Lee}]{lu2019vilbert}
Jiasen Lu, Dhruv Batra, Devi Parikh, and Stefan Lee. 2019.
\newblock Vilbert: Pretraining task-agnostic visiolinguistic representations
  for vision-and-language tasks.
\newblock \emph{Advances in Neural Information Processing Systems}, 32:13--23.

\bibitem[{Petroni et~al.(2019)Petroni, Rockt{\"a}schel, Riedel, Lewis, Bakhtin,
  Wu, and Miller}]{petroni2019language}
Fabio Petroni, Tim Rockt{\"a}schel, Sebastian Riedel, Patrick Lewis, Anton
  Bakhtin, Yuxiang Wu, and Alexander Miller. 2019.
\newblock Language models as knowledge bases?
\newblock In \emph{Proceedings of EMNLP-IJCNLP 2019}, pages 2463--2473.

\bibitem[{Qi et~al.(2020)Qi, Zhang, Zhang, Bolton, and Manning}]{qi2020stanza}
Peng Qi, Yuhao Zhang, Yuhui Zhang, Jason Bolton, and Christopher~D. Manning.
  2020.
\newblock Stanza: A {Python} natural language processing toolkit for many human
  languages.
\newblock In \emph{Proceedings of the 58th Annual Meeting of the Association
  for Computational Linguistics: System Demonstrations}.

\bibitem[{Qin and Eisner(2021)}]{qin2021learning}
Guanghui Qin and Jason Eisner. 2021.
\newblock Learning how to ask: Querying lms with mixtures of soft prompts.
\newblock In \emph{Proceedings of NAACL 2021}, pages 5203--5212.

\bibitem[{Radford et~al.(2021)Radford, Kim, Hallacy, Ramesh, Goh, Agarwal,
  Sastry, Askell, Mishkin, Clark et~al.}]{radford2021learning}
Alec Radford, Jong~Wook Kim, Chris Hallacy, Aditya Ramesh, Gabriel Goh,
  Sandhini Agarwal, Girish Sastry, Amanda Askell, Pamela Mishkin, Jack Clark,
  et~al. 2021.
\newblock Learning transferable visual models from natural language
  supervision.
\newblock \emph{arXiv preprint arXiv:2103.00020}.

\bibitem[{Radford et~al.(2018)Radford, Narasimhan, Salimans, and
  Sutskever}]{radford2018improving}
Alec Radford, Karthik Narasimhan, Tim Salimans, and Ilya Sutskever. 2018.
\newblock Improving language understanding by generative pre-training.

\bibitem[{Radford et~al.(2019)Radford, Wu, Child, Luan, Amodei, Sutskever
  et~al.}]{radford2019language}
Alec Radford, Jeffrey Wu, Rewon Child, David Luan, Dario Amodei, Ilya
  Sutskever, et~al. 2019.
\newblock Language models are unsupervised multitask learners.
\newblock \emph{OpenAI blog}, 1(8):9.

\bibitem[{Raffel et~al.(2020)Raffel, Shazeer, Roberts, Lee, Narang, Matena,
  Zhou, Li, and Liu}]{raffel2020exploring}
Colin Raffel, Noam Shazeer, Adam Roberts, Katherine Lee, Sharan Narang, Michael
  Matena, Yanqi Zhou, Wei Li, and Peter~J Liu. 2020.
\newblock Exploring the limits of transfer learning with a unified text-to-text
  transformer.
\newblock \emph{Journal of Machine Learning Research}, 21:1--67.

\bibitem[{Schick and Sch{\"u}tze(2021{\natexlab{a}})}]{schick2021exploiting}
Timo Schick and Hinrich Sch{\"u}tze. 2021{\natexlab{a}}.
\newblock Exploiting cloze-questions for few-shot text classification and
  natural language inference.
\newblock In \emph{Proceedings of the 16th Conference of the EACL}, pages
  255--269.

\bibitem[{Schick and Sch{\"u}tze(2021{\natexlab{b}})}]{schick2021s}
Timo Schick and Hinrich Sch{\"u}tze. 2021{\natexlab{b}}.
\newblock It’s not just size that matters: Small language models are also
  few-shot learners.
\newblock In \emph{Proceedings of the 2021 Conference of the North American
  Chapter of the Association for Computational Linguistics: Human Language
  Technologies}, pages 2339--2352.

\bibitem[{Sharma et~al.(2018)Sharma, Ding, Goodman, and
  Soricut}]{sharma2018conceptual}
Piyush Sharma, Nan Ding, Sebastian Goodman, and Radu Soricut. 2018.
\newblock Conceptual captions: A cleaned, hypernymed, image alt-text dataset
  for automatic image captioning.
\newblock In \emph{Proceedings of the 56th Annual Meeting of the Association
  for Computational Linguistics}, pages 2556--2565.

\bibitem[{Shen et~al.(2021)Shen, Li, Tan, Bansal, Rohrbach, Chang, Yao, and
  Keutzer}]{shen2021much}
Sheng Shen, Liunian~Harold Li, Hao Tan, Mohit Bansal, Anna Rohrbach, Kai-Wei
  Chang, Zhewei Yao, and Kurt Keutzer. 2021.
\newblock How much can clip benefit vision-and-language tasks?
\newblock \emph{arXiv preprint arXiv:2107.06383}.

\bibitem[{Shin et~al.(2020)Shin, Razeghi, Logan~IV, Wallace, and
  Singh}]{shin2020eliciting}
Taylor Shin, Yasaman Razeghi, Robert~L Logan~IV, Eric Wallace, and Sameer
  Singh. 2020.
\newblock Eliciting knowledge from language models using automatically
  generated prompts.
\newblock In \emph{Proceedings of EMNLP 2020}, pages 4222--4235.

\bibitem[{Su et~al.(2020)Su, Zhu, Cao, Li, Lu, Wei, and Dai}]{su2020vl}
Weijie Su, Xizhou Zhu, Yue Cao, Bin Li, Lewei Lu, Furu Wei, and Jifeng Dai.
  2020.
\newblock Vl-bert: Pre-training of generic visual-linguistic representations.
\newblock In \emph{International Conference on Learning Representations}.

\bibitem[{Tsimpoukelli et~al.(2021)Tsimpoukelli, Menick, Cabi, Eslami, Vinyals,
  and Hill}]{tsimpoukelli2021multimodal}
Maria Tsimpoukelli, Jacob Menick, Serkan Cabi, SM~Eslami, Oriol Vinyals, and
  Felix Hill. 2021.
\newblock Multimodal few-shot learning with frozen language models.
\newblock \emph{arXiv preprint arXiv:2106.13884}.

\bibitem[{Vaswani et~al.(2017)Vaswani, Shazeer, Parmar, Uszkoreit, Jones,
  Gomez, Kaiser, and Polosukhin}]{vaswani2017attention}
Ashish Vaswani, Noam Shazeer, Niki Parmar, Jakob Uszkoreit, Llion Jones,
  Aidan~N Gomez, {\L}ukasz Kaiser, and Illia Polosukhin. 2017.
\newblock Attention is all you need.
\newblock In \emph{Advances in neural information processing systems}, pages
  5998--6008.

\bibitem[{Wang et~al.(2021)Wang, Bao, Dong, and Wei}]{wang2021vlmo}
Wenhui Wang, Hangbo Bao, Li~Dong, and Furu Wei. 2021.
\newblock Vlmo: Unified vision-language pre-training with
  mixture-of-modality-experts.
\newblock \emph{arXiv preprint arXiv:2111.02358}.

\bibitem[{Wu and Dredze(2019)}]{wu2019beto}
Shijie Wu and Mark Dredze. 2019.
\newblock Beto, bentz, becas: The surprising cross-lingual effectiveness of
  bert.
\newblock In \emph{Proceedings of EMNLP 2019}, pages 833--844.

\bibitem[{Xie et~al.(2019)Xie, Lai, Doran, and Kadav}]{xie2019visual}
Ning Xie, Farley Lai, Derek Doran, and Asim Kadav. 2019.
\newblock Visual entailment: A novel task for fine-grained image understanding.
\newblock \emph{arXiv preprint arXiv:1901.06706}.

\bibitem[{Xue et~al.(2021)Xue, Constant, Roberts, Kale, Al-Rfou, Siddhant,
  Barua, and Raffel}]{xue2021mt5}
Linting Xue, Noah Constant, Adam Roberts, Mihir Kale, Rami Al-Rfou, Aditya
  Siddhant, Aditya Barua, and Colin Raffel. 2021.
\newblock mt5: A massively multilingual pre-trained text-to-text transformer.
\newblock In \emph{Proceedings of NAACL 2021}, pages 483--498.

\bibitem[{Young et~al.(2014)Young, Lai, Hodosh, and
  Hockenmaier}]{young2014image}
Peter Young, Alice Lai, Micah Hodosh, and Julia Hockenmaier. 2014.
\newblock From image descriptions to visual denotations: New similarity metrics
  for semantic inference over event descriptions.
\newblock \emph{Transactions of the Association for Computational Linguistics},
  2:67--78.

\bibitem[{Yu et~al.(2019)Yu, Yu, Cui, Tao, and Tian}]{yu2019deep}
Zhou Yu, Jun Yu, Yuhao Cui, Dacheng Tao, and Qi~Tian. 2019.
\newblock Deep modular co-attention networks for visual question answering.
\newblock In \emph{Proceedings of the IEEE/CVF Conference on Computer Vision
  and Pattern Recognition}, pages 6281--6290.

\bibitem[{Zhang et~al.(2021)Zhang, Li, Hu, Yang, Zhang, Wang, Choi, and
  Gao}]{zhang2021vinvl}
Pengchuan Zhang, Xiujun Li, Xiaowei Hu, Jianwei Yang, Lei Zhang, Lijuan Wang,
  Yejin Choi, and Jianfeng Gao. 2021.
\newblock Vinvl: Revisiting visual representations in vision-language models.
\newblock In \emph{Proceedings of the IEEE/CVF Conference on Computer Vision
  and Pattern Recognition}, pages 5579--5588.

\end{thebibliography}
\bibliographystyle{acl_natbib}

\newpage
\appendix

\section*{Appendix}

\section{Datasets Statistics}
\label{apdx:dataset}
\begin{table}[ht]
\small
\centering
\begin{tabular}{@{}l|rrr|r@{}}
\toprule
Datasets                  & \textit{\# Train} & \textit{\# Valid} & \textit{\# Test} & \textit{\# Vocab}       \\ \midrule
\multirow{2}{*}{VQAv2}   & 443,757           & 214,354           & -                & \multirow{2}{*}{19,174} \\
                         & 82,783            & 40,504            & -                &                         \\ \midrule
\multirow{2}{*}{SNLI-VE} & 529,527           & 17,858            & 17,901           & \multirow{2}{*}{32,191} \\
                         & 29,783            & 1,000             & 1,000            &                         \\ \bottomrule
\end{tabular}%
\caption{Basic statistics of the two datasets. The upper is the number of examples, and the lower is the number of distinct images. And \textit{\# Vocab} is the vocabulary size.}
\label{tab:datasets}
\end{table}

\section{Details of Implementation}
\subsection{Zero-shot Model Briefs}
In our experiments, we leverage two kinds of pre-trained models: the CLIP variants and the T5. We brief these models as follows.

For the CLIP models, the text encoder is always a transformer, but its hidden size varies according to the size of visual encoders. And there are two architectures of visual encoders, including the vision transformer (ViT) and ResNet.
\begin{itemize}
    \item \textbf{CLIP ViT-B/16}: both the text and visual encoders are 12-layer, 512-hidden transformers.
    \item \textbf{CLIP RN101}: the text encoder is a 12-layer transformer, and the visual encoder is ResNet101, both with a hidden size of 512.
    \item \textbf{CLIP RN50x16}: the text encoder is a 12-layer transformer, and the visual encoder is ResNet50x16, both with a hidden size of 768.
\end{itemize}
All CLIP models we used are from the official CLIP repository\footnote{https://github.com/openai/CLIP}.
For the language model T5, we use a publicly available T5$_\text{large}$ checkpoint from the Huggingface repository\footnote{https://huggingface.co/models}. The T5$_\text{large}$ has 24 hidden layers, 16 self-attention heads, 1024 hidden size, and a total of 770M parameters. It is trained on  Colossal Clean Crawled Corpus (C4).
Note that the T5 model had not been trained or finetuned under both few-shot and zero-shot settings.

\subsection{Hyperparameters}

\begin{table}[ht]
\centering
\small
\renewcommand\tabcolsep{3.5pt}
\begin{tabular}{lr}
\toprule
Hyperparameters & Value \\ \midrule
Training epochs & 30 \\
Batch size & 8 \\
Initial temperature & 0.07 \\
Maximum temperature & 100.0 \\
Adam $\epsilon$ & 1e-8 \\
Adam $\beta$ & (0.9, 0.999) \\
Learning rate & 2e-5 \\
Gradient clipping & 2.0 \\
Weight decay & 0.001 \\
Number of filtered answers & 200 \\
\bottomrule
\end{tabular}
\caption{Hyperparameters used for CLIP models in few-shot learning. 
}
\label{table:pt-hparam}
\end{table}

We report the hyperparameter settings of few-shot CLIP training in Table~\ref{table:pt-hparam}.
We apply the same set of hyperparameters to fine-tune both ResNet CLIP and ViT CLIP.

\begin{table}[ht]
\centering
\small
\renewcommand\tabcolsep{3.5pt}
\begin{tabular}{lr}
\toprule
Hyperparameters & Value \\ \midrule
Layers  & 3 \\
Layer Dimension & 1024-128-3 \\
Training epochs & 20 \\
Adam $\epsilon$ & 1e-8 \\
Adam $\beta$ & (0.9, 0.999) \\
Gradient clipping & 2.0 \\
Learning rate & \{1e-6, 3e-6, 5e-6\} \\
Batch size & \{32, 64, 128\} \\
Dropout & \{0, 0.1, 0.4\} \\
\bottomrule
\end{tabular}
\caption{Hyperparameters of the MLP classifier in zero-shot language to vision transfer.
}
\label{table:mlp-hparam}
\end{table}

The hyperparameters used for the MLP classifier in the visual entailment task are reported in Table~\ref{table:mlp-hparam}. We performed grid searches on the combination of the learning rate, batch size, and dropout.
The CLIP variants reached the best performances under different parameter combinations.

\begin{table}[ht]
\centering
\small
\renewcommand\tabcolsep{3.5pt}
\begin{tabular}{lr}
\toprule
\textbf{Template Generation}\\
Hyperparameters & Value \\ \midrule
Number of beams & 20 \\
Number of returned sequences & 10 \\
Max returned span length & 30 \\
\midrule
\textbf{Answer Filtering}\\
Hyperparameters  & Values \\
\midrule
Batch size & 128 \\
Number of beams & 200 \\
Number of returned sequences & 200 \\
Max returned span length & 6 \\
Max number of demonstration & 16\\
\bottomrule
\end{tabular}
\caption{Hyperparameters used for T5 in template generation and answer filtering.
}
\label{table:T5-hparam}
\end{table}

Table~\ref{table:T5-hparam} shows the hyperparameter configurations for T5's conditional generation, which is leveraged to generate the masked template and filter answers.

\subsection{The Number of Learnable Parameters}
\label{apdx:number-para}
\begin{table}[ht]
\centering
\small
\renewcommand\tabcolsep{3.5pt}
\begin{tabular}{@{}l|rrrr@{}}
\toprule
CLIP & \textit{\# Bias} & \textit{\# Normalize} & \textit{\# BiNor} & \textit{\# All} \\ \midrule
CLIP$_\text{RN101}$ & 127,488 & 123,392 & 189,184 & 100M \\
CLIP$_\text{RN50x16}$ & 209,088 & 132,160 & 319,488 & 229M \\
CLIP$_\text{ViT-B/16}$ & 171,008 & 65,536 & 203,776 & 149M \\ \bottomrule
\end{tabular}%
\caption{Statistics of different type of parameters in CLIP models.
}
\label{table:count-param}
\end{table}

Table~\ref{table:count-param} shows the number of different type of learnable parameter in CLIP models. The counting of Bias and Normalization share the $\beta$ in Eq.\ref{eq:binor}.
The numbers of BiNor parameters are about 0.2M to 0.3M, accounting for less than 0.3\% of all parameters.


\section{Few-shot Training Procedure}
\label{apdx:algorithm}
\renewcommand{\algorithmicrequire}{\textbf{Input:}}  
\renewcommand{\algorithmicensure}{\textbf{Output:}} 

\begin{algorithm}[ht]
\small
  \caption{CLIP models few-shot training.}
  \label{alg:fewshot_training}
  \begin{algorithmic}[1]
    \Require
      $\mathbb V$: visual encoder, ResNet or ViT;
      $\mathbb T$: text encoder, Transformer;
      $I$: few-shot images;
      $P$: few-shot prompts;
      $A$: few-shot answers;
      $\tau$: learned temperature parameter;
      $N$: max iterations;
      Adam: optimizer;
    \Ensure  Few-shot CLIP model.
    \State initial $epoch=0$, freeze parameters in $\mathbb V$ and $\mathbb T$ except bias and normalization;
    
    \Repeat{
      \State Sample C-way K-shot $E$ from ($I$,$P$,$A$);
      \State Split $E$ into \textit{support set} and \textit{query set};
      \ForAll {minibatch (i,p,a) in \textit{support set}}
          \State I$_{f}$ = $\mathbb V$(i), T$_{f}$ = $\mathbb T$(p);
          \State I$_{e}$ = norm(I$_{f}$), T$_{e}$ = norm(T$_{f}$);
          \State logits = $\tau$ * dot($I_{e}$, $T_{e}$);
          \State labels = map(a, p);
          \State loss = cross\_entropy(logits, labels);
          \State Adam.step();
      \EndFor;
      \State $epoch = epoch + 1$;
      \State Evaluate on \textit{query set};
    \Until{($epoch \ge N$)}.
    }
  \end{algorithmic}
\end{algorithm}

\section{Examples of Template Generation}
\label{apdx:template-generation}

In this section, we showcase several template generation examples to illustrate how the proposed method works. 
Since we have introduced how to convert a question into a masked template by demonstrating examples to the T5 (\cref{sec:t_vqa}), here we directly present several examples in Table~\ref{tab:template-generation}. These examples are sampled from five different question types and also cover the three answer types. 
As shown in Table~\ref{tab:template-generation}, a single demo in the demonstration consists of a question and an answer with the [mask] token. Notice that the [mask] token is only a placeholder rather than a real mask in the pre-trained language models. Different from the \textit{<extra\_id\_0>} in T5 that represents a corrupted span, the [mask] is used to inform the T5 where the answer words should be placed. After seeing several examples in the demonstration, the powerful T5$_{\text{large}}$ model could capture the conversion pattern in each type of question and perfectly complete most conversions without ignoring the subtle grammars. Once the masked template is generated, we could infill the [mask] place with answer words and then carry out further processing. The processing for the \textit{yes/no} type is a little different: as it is a binary task, we directly generate a positive prompt and a negative prompt, rather than masked templates, for the \textit{yes} and \textit{no}, respectively.

\begin{table*}[ht]
\centering
\resizebox{\textwidth}{!}{%
\begin{tabular}{@{}ll@{}}
\toprule
\multicolumn{1}{l|}{\textbf{Question Type}} & what color is \\ \midrule
\multicolumn{1}{l|}{\textbf{Demonstration}} & \begin{tabular}[c]{@{}l@{}}What color is the floor of this area? The color of floor of this area is {[}mask{]}.\\ What color is the pillow the cat is on? The color of the pillow the cat on is {[}mask{]}.\\ What color is the child's shorts? The color of the child's shorts is {[}mask{]}.\\ What color is the lettering on the business sign? The color of the lettering on the business sign is {[}mask{]}.\end{tabular} \\ \midrule
\multicolumn{1}{l|}{\textbf{Question}} & What color is the fence behind the man? \\
\multicolumn{1}{l|}{\textbf{Generated Template}} & The color of the fence behind the man is {[}mask{]}. \\ \midrule
\multicolumn{1}{l|}{\textbf{Question}} & What color is the statue near the building? \\
\multicolumn{1}{l|}{\textbf{Generated Template}} & The color of the statue near the building is {[}mask{]}. \\ \midrule
\textbf{} &  \\ \midrule
\multicolumn{1}{l|}{\textbf{Question Type}} & why is the \\ \midrule
\multicolumn{1}{l|}{\textbf{Demonstration}} & \begin{tabular}[c]{@{}l@{}}Why is the ground surface near the train a different color? The ground surface near the train is in a different color because of {[}mask{]}.\\ Why is the cat under an umbrella? The cat under an umbrella is because of {[}mask{]}.\\ Why is the laptop sitting above a larger keyboard? The laptop is sitting above a larger keyboard because of {[}mask{]}.\\ Why is the car being towed? The car is being towed because of {[}mask{]}.\end{tabular} \\ \midrule
\multicolumn{1}{l|}{\textbf{Question}} & Why is the little boy having fun? \\
\multicolumn{1}{l|}{\textbf{Generated Template}} & The little boy is having fun because of {[}mask{]}. \\ \midrule
\multicolumn{1}{l|}{\textbf{Question}} & Why is the elephant's trunk two color's? \\
\multicolumn{1}{l|}{\textbf{Generated Template}} & The elephant's trunk is two colors because of {[}mask{]}. \\ \midrule
\textbf{} &  \\ \midrule
\multicolumn{1}{l|}{\textbf{Question Type}} & which \\ \midrule
\multicolumn{1}{l|}{\textbf{Demonstration}} & \begin{tabular}[c]{@{}l@{}}Which utensil is on the table in the foreground? The {[}mask{]} utensil is on the table in the foreground.\\ Which way is the train going? The {[}mask{]} way is the train going.\\ Which hand holds the racket? The {[}mask{]} hand holds the racket.\\ Which foot is lifted in the air? The {[}mask{]} foot is lifted in the air.\end{tabular} \\ \midrule
\multicolumn{1}{l|}{\textbf{Question}} & Which hot dog has a larger variety of toppings? \\
\multicolumn{1}{l|}{\textbf{Generated Template}} & The {[}mask{]} hot dog has a larger variety of toppings. \\ \midrule
\multicolumn{1}{l|}{\textbf{Question}} & Which operating system is being used on this computer? \\
\multicolumn{1}{l|}{\textbf{Generated Template}} & The {[}mask{]} operating system is being used on this computer. \\ \midrule
\multicolumn{1}{l|}{\textbf{Question}} & Which side of the room is the television probably on? \\
\multicolumn{1}{l|}{\textbf{Generated Template}} & The {[}mask{]} side of the room is the television probably on. \\ \midrule
\textbf{} &  \\ \midrule
\multicolumn{1}{l|}{\textbf{Question Type}} & how many \\ \midrule
\multicolumn{1}{l|}{\textbf{Demonstration}} & \begin{tabular}[c]{@{}l@{}}How many unopened rolls of paper are in the picture? There are {[}mask{]} unopened rolls of paper in the picture.\\ How many engines does the closest airplane have? The closest airplane has {[}mask{]} engines.\\ How many different types of doors are visible? There are {[}mask{]} different types of doors visible.\\ How many people are wearing plaid shirts? There are {[}mask{]} people wearing plaid shirts.\end{tabular} \\ \midrule
\multicolumn{1}{l|}{\textbf{Question}} & How many people are participating in the eating contest? \\
\multicolumn{1}{l|}{\textbf{Generated Template}} & There are {[}mask{]} people participating in the eating contest. \\ \midrule
\multicolumn{1}{l|}{\textbf{Question}} & How many cabinets have been installed? \\
\multicolumn{1}{l|}{\textbf{Generated Template}} & There are {[}mask{]} cabinets installed. \\ \midrule
\multicolumn{1}{l|}{\textbf{Question}} & How many people in this picture are wearing glasses? \\
\multicolumn{1}{l|}{\textbf{Generated Template}} & There are {[}mask{]} people wearing glasses. \\ \midrule
\textbf{} &  \\ \midrule
\multicolumn{1}{l|}{\textbf{Question Type}} & does this \\ \midrule
\multicolumn{1}{l|}{\textbf{Demonstration}} & \begin{tabular}[c]{@{}l@{}}Positive:\\     Does this food look burnt? This food looks burnt.\\     Does this appear to be a noisy environment? This appears to be a noisy environment.\\      \\ Negative:\\     Does this boat have an engine? This boat has no engine.\\     Does this type of fruit change color? This type of fruit does not change color.\\ 
Does this animal produce dairy products? This animal does not produce dairy products.\\     Does this pizza look hot? This pizza does not look hot.\end{tabular} \\ \midrule
\multicolumn{1}{l|}{\textbf{Question}} & Does this look like a happy occasion? \\
\multicolumn{1}{l|}{\textbf{Generated Prompts}} & \begin{tabular}[c]{@{}l@{}}Yes$\rightarrow$ This looks like a happy occasion\\ No $\rightarrow$ This does not look like a happy occasion\end{tabular} \\ \midrule
\multicolumn{1}{l|}{\textbf{Question}} & Does this man have both of his skis on? \\
\multicolumn{1}{l|}{\textbf{Generated Prompts}} & \begin{tabular}[c]{@{}l@{}}Yes$\rightarrow$ This man has both of his skis on\\ No $\rightarrow$ This man does not have both of his skis on\end{tabular} \\ \midrule
\multicolumn{1}{l|}{\textbf{Question}} & Does this transportation run on gasoline? \\
\multicolumn{1}{l|}{\textbf{Generated Prompts}} & \begin{tabular}[c]{@{}l@{}}Yes$\rightarrow$ This transportation runs on gasoline\\ No $\rightarrow$ This transportation does not run on gasoline\end{tabular} \\ \bottomrule
\end{tabular}%
}
\caption{Examples of generating masked templates. The demonstrations are defined for each type of question and are demonstrated to the T5. For the binary \textit{yes/no} type, we directly generate positive prompt for \textit{yes} and negative prompt for \textit{no}. Full question types are available at https://github.com/GT-Vision-Lab/VQA/tree/master/QuestionTypes.}
\label{tab:template-generation}
\end{table*}

\end{document}